\documentclass{article}

\PassOptionsToPackage{sort, compress}{natbib}

\usepackage[preprint]{neurips_2026}

\usepackage[utf8]{inputenc} 
\usepackage[T1]{fontenc}    
\usepackage{hyperref}       
\usepackage{url}            
\usepackage{booktabs}       
\usepackage{amsfonts}       
\usepackage{nicefrac}       
\usepackage{microtype}      
\usepackage{xcolor}         

\usepackage{amsmath}
\usepackage{amssymb}
\usepackage{mathtools}
\usepackage{amsthm}
\usepackage{graphicx}
\usepackage[ruled,vlined]{algorithm2e} 
\usepackage[T1]{fontenc}
\usepackage{subcaption}  
\usepackage{arydshln} 
\usepackage{tabularx}
\usepackage{units}
\usepackage[inline]{enumitem}
\usepackage[capitalize,noabbrev]{cleveref}

\newcolumntype{L}{>{\hsize=1.6\hsize\raggedright\arraybackslash}X} 
\newcolumntype{R}{>{\hsize=1.0\hsize\raggedright\arraybackslash}X} 

\usepackage[acronym,nonumberlist,nogroupskip,nopostdot]{glossaries}
\makeglossaries
\glsdisablehyper

\newacronym{llm}{LLM}{large language model}
\newacronym{mlp}{MLP}{multilayer perceptron}
\newacronym{mha}{MHA}{multi-head attention}
\newacronym{ebm}{EBM}{energy-based model}
\newacronym{cem}{CEM}{Causal Energy Minimization}
\newacronym{cccp}{CCCP}{Concave-Convex Procedure}

\newcommand{\cem}[1]{\textcolor{orange!85!black}{#1}}

\theoremstyle{plain}

\theoremstyle{definition}

\theoremstyle{remark}

\usepackage{amsmath,amsfonts,bm}

\newcommand{\newterm}[1]{{\bf #1}}

\def\eqref#1{equation~\ref{#1}}

\def\1{\bm{1}}

\def\vzero{{\bm{0}}}

\def\va{{\bm{a}}}

\def\vd{{\bm{d}}}

\def\vg{{\bm{g}}}
\def\vh{{\bm{h}}}

\def\vk{{\bm{k}}}

\def\vp{{\bm{p}}}
\def\vq{{\bm{q}}}

\def\vu{{\bm{u}}}
\def\vv{{\bm{v}}}

\def\vx{{\bm{x}}}

\def\vbeta{{\bm{\beta}}}
\def\vgamma{{\bm{\gamma}}}

\def\mA{{\bm{A}}}

\def\mD{{\bm{D}}}

\def\mI{{\bm{I}}}

\def\mP{{\bm{P}}}

\def\mR{{\bm{R}}}

\def\mU{{\bm{U}}}
\def\mV{{\bm{V}}}
\def\mW{{\bm{W}}}

\DeclareMathAlphabet{\mathsfit}{\encodingdefault}{\sfdefault}{m}{sl}
\SetMathAlphabet{\mathsfit}{bold}{\encodingdefault}{\sfdefault}{bx}{n}

\def\sR{{\mathbb{R}}}

\newcommand{\diag}{\operatorname{diag}}

\newcommand{\R}{\mathbb{R}}

\newcommand{\softmax}{\operatorname{softmax}}
\newcommand{\gmlp}{\operatorname{GatedMLP}}
\newcommand{\rms}{\operatorname{RMSNorm}}
\newcommand{\mha}{\operatorname{MHA}}

\newcommand{\softplus}{\operatorname{softplus}}

\DeclareMathOperator{\sign}{sign}

\title{Revisiting Transformer Layer Parameterization Through Causal Energy Minimization}

\author{%
  Jin Xu\thanks{Correspondence to \texttt{jinxu2@microsoft.com} and \texttt{jameshensman@microsoft.com}} \\
  Microsoft
  \And
  Camille Couturier \\
  Microsoft
  \And
  Victor Rühle \\
  Microsoft
  \AND
  Saravan Rajmohan \\
  Microsoft
  \And
  James Hensman\footnotemark[1] \\
  Microsoft Research Cambridge
}

\begin{document}

\maketitle

\begin{abstract}
Transformer blocks typically combine multi-head attention (MHA) for token mixing with gated MLPs for token-wise feature transformation, yet many choices in their parameterization remain largely empirical. We introduce Causal Energy Minimization (CEM), a framework that recasts Transformer layers as optimization steps on conditional energy functions while explicitly accounting for layer parameterization. Extending prior energy-based interpretations of attention, CEM shows that weight-tied MHA can be derived as a gradient update on an interaction energy, and that a gated MLP with shared up/down projections can be viewed through an element-wise energy. This perspective identifies a design space for Transformer layers that includes within-layer weight sharing, diagonal-plus-low-rank interactions, lightweight preconditioners, and recursive updates. We evaluate CEM-derived layers in language-modeling experiments at the moderate hundred-million-parameter scale. Despite their constrained parameterizations, these layers train stably and can match corresponding Transformer baselines. Overall, our results suggest that CEM provides a useful lens for understanding Transformer layer parameterization, connecting Transformer architectures to energy-based models and motivating further exploration of energy-guided layer designs.
\end{abstract}

\section{Introduction}
Stacked sequence-to-sequence mappings underlie modern foundation models \citep{bommasani2021opportunities}. Early work employed recurrent \citep{sutskever2014sequence,hochreiter1997long,cho2014rnmt} and convolutional architectures \citep{kalchbrenner2016neural,gehring2017convolutional}, but these have been largely replaced by Transformer \citep{vaswani2017attention}. Despite alternatives such as structured state-space models \citep{gu2021efficiently,gu2023mamba}, Transformer layers, particularly \glspl{mha} and gated \glspl{mlp}, remain the core building blocks of today’s \glspl{llm}. Yet architectural innovations for Transformers continue to be driven mainly by empirical studies \citep{shazeer2017moe,shazeer2020glu,shazeer2019fast,ainslie2023gqa}, motivating perspectives that connect these parameterizations to explicit computational principles.

\Glspl{ebm} provide one such perspective. By assigning a
scalar energy $\mathcal{E}(\vx)$ to a configuration $\vx$, \glspl{ebm} interpret
computation as the search for low-energy states
\citep{lecun2006tutorial,hopfield1982neural,ackley1985learning,krotov2016dense,ramsauer2021hopfield}.
This viewpoint connects layer computation to optimization, making it possible to
study architecture and parameterization through energy functions, update rules, and
optimization dynamics. 

We introduce \acrfull{cem}, a framework that associates Transformer
layers with conditional energy functions and interprets their residual
updates as optimization steps. Concretely, to map an input sequence
$\vh_{1:J}$ to an output sequence $\vh'_{1:J}$, \gls{cem} introduces, for each
position $i$, an optimization variable $\vx_i$ initialized at $\vh_i$. The
variable is updated by a procedure $\mathcal{A}$ using a conditional energy
$\epsilon(\vx_i \mid \vh_{1:i})$ that depends on the causal history
$\vh_{1:i}$, and the resulting state is used as the output $\vh'_i$. 
This formulation separates the roles of the energy and the update
procedure, providing a unified way to analyze and design layer
transformations.

Building on prior energy-based views of attention \citep{ramsauer2021hopfield,sander2022sinkformers}, \gls{cem} focuses on the
parameterization induced by the energy-gradient perspective. We show that weight-tied \gls{mha} arises
as such a step on an interaction energy, with the key projection tied to the
value projection and the query projection tied to the output projection, and
that gated \glspl{mlp} with shared up/down projections admit an analogous
interpretation through element-wise energies
(\Cref{subsec:attention-as-gradient,subsec:mlp-as-gradient}). Our goal is not
to replace Transformers with standalone energy-based sequence models, which
have faced challenges on large-scale language modeling
\citep{du2021improved,qin2022cold}, but to ask whether Transformer layers
themselves can be revisited as energy-based updates and what this perspective
implies for their parameterization and extension.

\begin{figure*}[ht!]
\begin{center}
\includegraphics[width=0.9\textwidth]{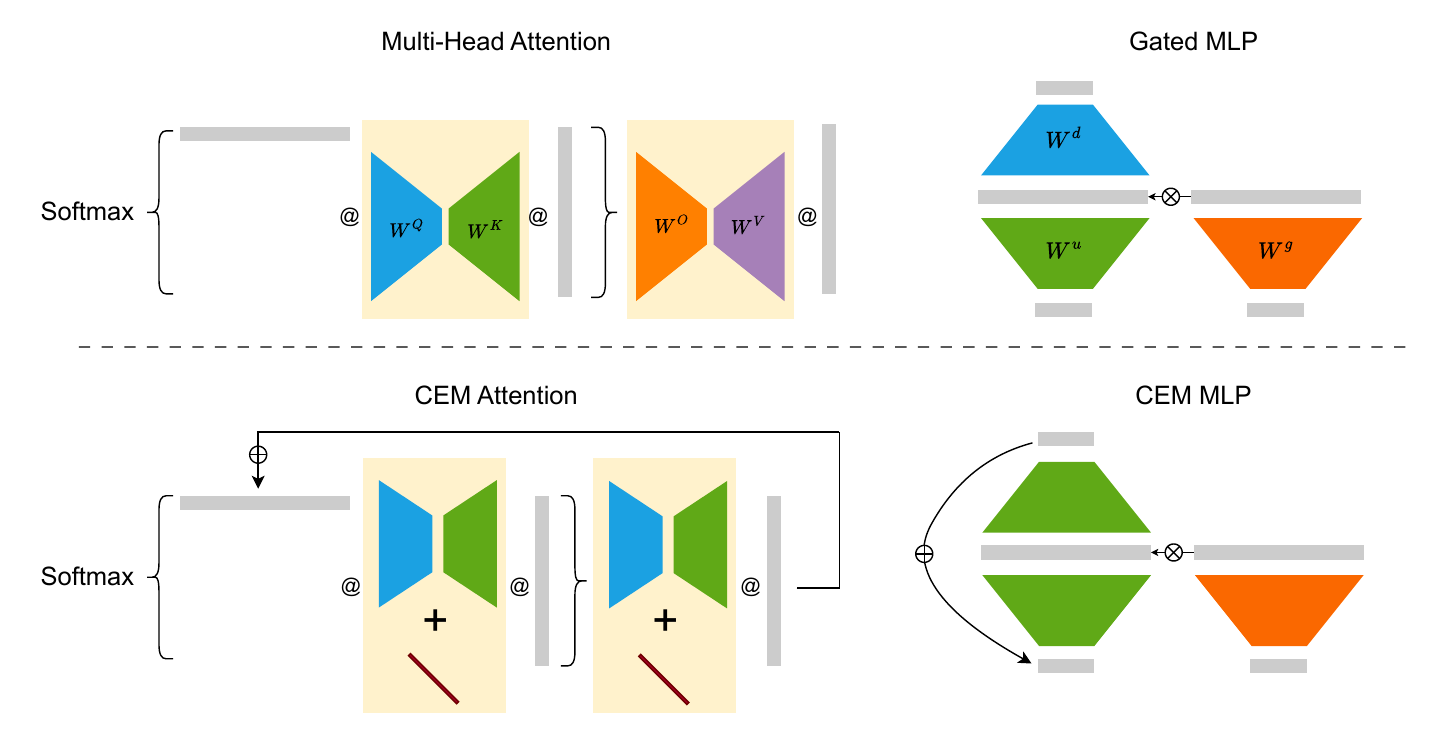} 
\end{center}
\caption{\textbf{Comparison of transformer layer parameterizations}.  
Top left: standard multi-head attention (per head). 
Top right: gated \gls{mlp}.  
Bottom left: \gls{cem}-derived attention.  
Bottom right: \gls{cem}-derived \gls{mlp}.  
Colors indicate shared weights within each subfigures (See \Cref{subsec:attention-as-gradient,subsec:mlp-as-gradient}). Arrows highlight the recursive structure of \gls{cem} modules, which implement multiple gradient steps of energy minimization (\Cref{eq:attn-recursion,eq:mlp_recursion}), while brown bars denote the optional diagonal term added to the key–query projections (See details in \Cref{subsec:enhance_layer_optimization_perspective}).
For the attention heads, $W^V$ maps the hidden state to values, which are then projected back by $W^O$ and scaled by the scalar Softmax weight.
}
\label{fig:cem_illustration}
\end{figure*}

\textbf{Contributions.}
We use \gls{cem} to study Transformer layer
parameterization, focusing on how energy functions and optimization updates can
inform the design of layer variants. Our experiments
evaluate these ideas at moderate scale and provide evidence that CEM-derived
layers can match standard Transformer components while enabling new
parameterization choices. In particular,
\begin{itemize}[leftmargin=1.5em,topsep=0pt, parsep=0pt]
    \item We connect shared key/query and value/output projections in \gls{mha},
    and shared up/down projections in \gls{mlp}, to gradient updates on
    interaction and element-wise energies, respectively.
    \item We extend layer design beyond single gradient updates, investigating
    diagonal-plus-low-rank weights, preconditioned updates, and within-layer
    recursive steps.
    \item We show that \gls{cem}-derived layers match their Llama counterparts
    at moderate scale, with parameterization choices that improve performance
    without increasing parameter count.
\end{itemize}

\section{Transformer layers as energy updates} \label{sec:transformer_layer_as_energy_updates}

We reframe Transformer layers through the lens of \gls{cem}. While prior work has explored energy-based views of attention \citep{ramsauer2021hopfield,sander2022sinkformers}, we extend this perspective to \glspl{mlp} and uncover a common weight-sharing structure across both components. We introduce two complementary energy terms: an \emph{interaction term}, capturing dependencies across features of different tokens, and an \emph{element-wise term}, assigning energy to each token's feature vector. Gradient-based updates on these energies naturally recover standard Transformer layers with weight sharing, shared key/value and query/output projections in \gls{mha} emerge from interaction-energy updates, while shared up/down projections in \glspl{mlp} emerge from element-wise energy updates. \Cref{fig:cem_illustration} illustrates the resulting scheme.

\subsection{Gradient of interaction energy yields weight-tied attention} 
\label{subsec:attention-as-gradient}

\paragraph{Multi-head attention (MHA).}
In conventional \gls{mha}, the query, key, and value projections for head $k$ are defined as
\[
    \vq_i^k = \mW_k^Q \vh_i, 
    \quad \vk_j^k = \mW_k^K \vh_j, 
    \quad \vv_j^k = \mW_k^V \vh_j,
\]
where $\vh_j$ denotes $j$-th token feature vector in the sequence.
The attention update then takes the form
\begin{align}
\mha(\vh_{1:i})
&= \sum_{k=1}^K \mW_k^{O\top}\!\left(\sum_{j=1}^i \alpha_{ij}^k\,\vv_j^k\right), \;
\text{where } \; \alpha_{ij}^k
= \softmax_j\!\left(\left\{\tfrac{1}{\sqrt{D_r}}(\vk_{j'}^k)^\top \vq_i^k\right\}_{j'=1}^i\right). 
\end{align}

Typically, the per-head outputs are concatenated and followed by a single output projection $\mW^O$. Equivalently, one may view $\mW^O$ as partitioned into head-specific blocks $\{\mW_k^O\in \mathbb{R}^{D_r \times D_h} \}_{k=1}^K$, with contributions summed as written above, where $D_h$ is the feature dimension for $\vh_i$ and $D_r$ is the head dimension where $\vq_i^k,\vk_j^k,\vv_j^k \in \mathbb{R}^{D_r}$. See $\Cref{subsec:concatenation_summation}$ for detailed explanation.
\paragraph{Interaction energy.} 
\Gls{mha} can be derived by considering a gradient step on the following simple interaction energy, similar to that in modern Hopfield networks \citep{ramsauer2021hopfield}:
\begin{align} \label{eq:energy_interaction}
    \epsilon(\vx_i \mid \vh_{1:i}) 
    &= -\tau \sum_{k=1}^K \log\sum_{j=1}^i 
        \exp\!\Big( \tfrac{1}{\tau}\,\vbeta_{kj}^{\top}\vx_i\Big) \quad \text{where } \quad  \vbeta_{kj} = \mA_k \vh_j. 
\end{align}
Here $\{\mA_k \in \sR^{D_h \times D_h}\}_{k=1}^K$ are learnable projection matrices, $D_h$ is the feature dimension, and $\tau$ is a scalar temperature. Our formulation differs from Hopfield networks in two respects: projection weights are embedded directly in the energy and reappear as tied attention projections, and we perform gradient updates rather than \gls{cccp} iterations (see \Cref{appsec:hopfield}).
We now derive the gradient of the interaction energy $\epsilon(\vx_i \mid \vh_{1:i})$ with respect to $\vx_i$:
\begin{align} \label{eq:attn-gradient-head}
    &\nabla_{\vx_i}\, \epsilon(\vx_i \mid \vh_{1:i}) = -\sum_{k=1}^{K} \sum_{j=1}^i 
    \softmax_j \Big( \big\{\tfrac{1}{\tau}\,\vbeta_{kj'}^\top \vx_i \big\}_{j'=1}^{i} \Big) \vbeta_{kj}\,.
\end{align}
Adopting a low-rank factorization $\mA_k = \mW_k^{Q\top}\mW_k^K$ 
we obtain $\vbeta_{kj} = \mW_k^{Q\top} (\mW_k^K \vh_j)$.
If our chosen algorithm is to take a single gradient step, initialized at  $\vx_i=\vh_i$, then we compute:
\begin{align}
    \vh'_i &= \vh_i -\eta_{\epsilon} \, \nabla_{\vx_i}\, \epsilon(\vx_i \mid \vh_{1:i})\Big|_{\vx_i=\vh_i}\,,
\end{align}
with
\begin{align} 
    \nabla_{\vx_i}\, \epsilon(\vx_i \mid \vh_{1:i}) \Big|_{\vx_i=\vh_i}
    = -\sum_{k=1}^{K} \mW_k^{Q\top} \Bigg(\sum_{j=1}^{i} 
        \alpha_{ij}^k \, \vv_j^k \Bigg) \nonumber \\\text{where } \quad \alpha_{ij}^k = \softmax_j\!( \big\{\tfrac{1}{\tau} (\vk_{j'}^k)^\top \vq_i^k \big\}_{j'=1}^i) \nonumber
\end{align}

where the value and key projections are shared: $\vq_i^k = \mW_k^Q \vh_i\,,\;\vv_j^k = \vk_j^k = \mW_k^K \vh_j$.

It is therefore clear that the gradient of the interaction energy recovers the \gls{mha} form,  
with the weight-tied parameterization \begin{align} \label{eq:attn_weights_sharing}
    \mW_k^K = \mW_k^V \qquad \mW_k^Q = \mW_k^O \qquad \tau = \sqrt{D_r} \,.
\end{align}
In this view, the residual update corresponds exactly to a single gradient descent step (with step size $\eta^{\epsilon}=1$) on the defined interaction energy.
\subsection{Gradient of element-wise energy yields weight-tied MLPs} \label{subsec:mlp-as-gradient}
\paragraph{Gated \glspl{mlp}.} 
A gated \gls{mlp} applies an element-wise transformation to the hidden state $\vh_i \in \R^{D_h}$:
\begin{equation}
\label{eq:gated_mlp}
    \gmlp(\vh_i) 
    = \mW^{\text{d}}\!\big( (\mW^{g}\vh_i) 
    \circ \sigma(\mW^{u}\vh_i) \big).
\end{equation}
Here, the learnable parameters are the \emph{gate} and \emph{up} projections $\mW^{g}, \mW^{u} \in \R^{D_m \times D_h}$ and the \emph{down} projection $\mW^{\text{d}} \in \R^{D_h \times D_m}$.  
The function $\sigma$ denotes a pointwise nonlinearity (e.g. GELU).
\paragraph{Element-wise energy term.} 
This energy term assigns energy independently to each token feature vector, while sharing the same functional form across positions:
\begin{equation} \label{eq:energy_elementwise}
    \xi(\vx_i \mid \vh_i) 
    = -\vgamma_i^\top \phi(\mV \vx_i),
    \quad \text{where }\quad \vgamma_i = \mW \vh_i\,.
\end{equation}
Here, the learnable parameters are the projection matrices $\mW, \mV \in \R^{D_v \times D_h}$, with projection dimension $D_v$ not necessarily equal to the hidden dimension $D_h$.  
The function $\phi$ denotes a pointwise nonlinearity.
\paragraph{Energy-gradient formulation.}
For the element-wise energy $\xi$,
the gradient with respect to $\vx_i$ is
\begin{align}
    \nabla_{\vx_i}\, \xi(\vx_i \mid \vh_i) 
    &= -\mV^\top \big(\vgamma_i \circ \phi'(\mV \vx_i)\big) \,.
\end{align}
Taking one gradient step at $\vx_i=\vh_i$ yields \begin{align}
    \vh'_i = \vh_i - \eta_{\xi} \, \nabla_{\vx_i}\, \xi(\vx_i \mid \vh_i)\Big|_{\vx_i=\vh_i}\,,
\end{align}
with
\begin{align} \label{eq:elemwise_gradient_eval}
     &\nabla_{\vx_i}\, \xi(\vx_i \mid \vh_i)\Big|_{\vx_i=\vh_i} = -\mV^\top \big((\mW \vh_i) \circ \phi'(\mV \vh_i)\big)\,.
\end{align}
Comparing \eqref{eq:gated_mlp} and \eqref{eq:elemwise_gradient_eval}, the energy-gradient update recovers the structure of gated \glspl{mlp} when we identify the parameters as
\begin{equation} \label{eq:mlp_weights_sharing}
   \mW^{d\top} = \mW^u = \mV\,,
    \qquad \mW^g = \mW\,,
    \qquad 
\end{equation}
and we set $\phi(x) = \int_{-\infty}^x \sigma(z) \text{d}z$.

\subsection{Enhancing transformer layers from an energy optimization perspective} \label{subsec:enhance_layer_optimization_perspective}
Having shown that Transformer layers, both MHA and \glspl{mlp}, can be interpreted as gradient updates on energy functions in \Cref{subsec:attention-as-gradient,subsec:mlp-as-gradient}, we next explore how these layers can be enhanced from the perspective of energy optimization. 
\paragraph{Diagonal-plus-low-rank parameterization}.  
In \Cref{subsec:attention-as-gradient}, we introduced a low-rank parameterization of $\mA_k=\mW_k^{Q\top}\mW_k^K$ in the interaction energy, recovering the query and key projections of standard attention. We now ask whether a purely low-rank form is sufficient, and instead propose a diagonal-plus-low-rank parameterization for the matrix $\mA_k$:  
\begin{align}\label{eq:diag_plus_low_rank_kv}
    \mA_k \;=\; \diag(\vd_k) \;+\; \mW_k^{Q\top} \mW_k^K ,
\end{align}
where $\vd_k \in \R^{D_h}$.  
This augmented parameterization captures key–query interactions that low-rank matrices alone cannot represent, yielding a richer structure for the interaction matrix $\mA_k$. The diagonal term enriches the interaction matrix but increases computational cost, so we propose sharing it across heads. A detailed empirical analysis is provided in \Cref{fig:kq_diag} and more background on this parameterization can be found in \Cref{diag-plus-low-rank-parameterization}.
\paragraph{Learned lightweight preconditioners.}  
A single gradient step may be insufficient to produce a well-optimized update.
Second-order methods such as Newton's method improve optimization by rescaling
gradients with curvature information, but computing such curvature is
prohibitively expensive in Transformer layers. Inspired by this idea, we
introduce lightweight learned preconditioner matrices with a
diagonal-plus-low-rank structure,
\begin{align} \label{eq:diag_plus_low_rank_preconditioners}
     \mP = \diag\!\big(\softplus(\vd)\big) + \mU\mV^\top + \mV\mU^\top,
\end{align}
where $\vd \in \R^{D_h}$ and $\mU,\mV \in \R^{D_h \times R}$ with $R \ll D_h$.
The positive diagonal term provides a stable base scaling, while the symmetric
low-rank correction captures richer curvature information at negligible cost.
We make no claim that $\mP$ approximates the true Hessian; rather, we treat it
as a learned proxy that can capture useful curvature information.

For the interaction energy, we insert per-head matrices $\mP_k$, giving
\begin{align}
    \Delta \vx_i^{\epsilon}(\vh_i \mid \vh_{1:i})
    \coloneqq - \sum_{k=1}^K \mP_k\, \mW_k^{Q\top}
        \Bigg(\sum_{j=1}^i
            \alpha_{ij}^k \vv_j^k \Bigg)\,, \\
    \text{where} \quad
    \alpha_{ij}^k =
    \softmax_j\!\Big(
        \big\{\tfrac{1}{\tau}(\vk_{j'}^k)^\top \vq_i^k\big\}_{j'=1}^{i}
    \Big) . \nonumber
\end{align}
This denotes the update evaluated at $\vx_i=\vh_i$ for the interaction energy
$\epsilon$.

For the element-wise energy, the gated \gls{mlp} update becomes (contrast with unpreconditioned one in \Cref{eq:elemwise_gradient_eval}):
\begin{align}
    &\Delta \vx_i^{\xi}(\vh_i \mid \vh_{1:i}) \coloneqq - \mP_{\operatorname{mlp}} \mV^\top \big((\mW \vh_i) \circ \phi'(\mV \vh_i)\big)\,,
\end{align}
with $\mP_{\operatorname{mlp}}$ denoting its preconditioner.  

In both cases, the preconditioners could be trained to provide lightweight curvature information, enabling updates that converge more effectively to well-optimized states. 
\paragraph{Multiple recursive steps}  
So far, each Transformer layer has been interpreted as performing a single gradient step on its associated energy function.  
From the optimization viewpoint, however, a single step rarely reaches a well-optimized state.  
A natural extension is therefore to apply multiple recursive updates within the same layer, analogous to running several iterations of an optimization algorithm.  
For the interaction energy (attention), starting from $\vx_i^{(0)} = \vh_i$, we perform $T$ updates of the form  
\begin{align} \label{eq:attn-recursion}
    \vx_i^{(t+1)} 
    = \vx_i^{(t)} - \eta_{\epsilon}\,\Delta \vx_i^{\epsilon}(\vx_i \mid \vh_{1:i}),
\end{align}
for $t = 0, \dots, T-1$ and set $\vh'_i = \vx_i^{(T)}$.  

For the element-wise energy (MLP), starting from $\vx_i^{(0)} = \vh_i$, the recursion is  
\begin{align} \label{eq:mlp_recursion}
    \vx_i^{(t+1)} 
    = \vx_i^{(t)} - \eta_{\xi}\, \Delta \vx_i^{\xi}(\vx_i \mid \vh_{1:i}),
\end{align}
for $t = 0, \dots, T-1$, with the output $\vh'_i = \vx_i^{(T)}$.  
This recursive scheme enables each layer to better minimize its energy function without adding parameters, as illustrated in \Cref{fig:cem_illustration}. Unlike blockwise recursion in looped Transformer, our approach updates only $\vx_i^{(t)}$ and fix $\vh_{1:i}$, with most computation performed outside the recursion. This within-layer recursion thus offers a distinct mechanism that could provide a new dimension for test-time scaling, which we leave for future work.

\subsection{A construction of transformer block with energy updates}
We now present the full Transformer block from the \gls{cem} perspective, where both attention and \gls{mlp} components arise as recursive gradient updates on their respective energy functions. Residual connections are absorbed into the recursion, while $\rms(\cdot)$ are applied. Standard Transformer with weight sharing, as detailed in \Cref{eq:attn_weights_sharing,eq:mlp_weights_sharing}, appears as the special case $T_\epsilon=T_\xi=1$, using identity preconditioners $(\mP_k)=(\mP_{\operatorname{mlp}})=\mI$ and vanishing diagonal terms $\vd_k=\vzero$. The complete \gls{cem} block is summarized in \Cref{algo:cem_block}.

\begin{algorithm*}
\centering
\small
\parbox[t]{\linewidth}{
\textbf{Input:}
Sequence $\vh_{1:J}$; \\
\textbf{Output:} Sequence $\vh'_{1:J}$.\\
\textbf{Hyperparameters:} Recursive steps $T_\epsilon,T_\xi$, step sizes $\eta_{\epsilon},\eta_{\xi}$, heads $K$, 
$\phi(x)=\int_{-\infty}^x \operatorname{SiLU}(z)\,\mathrm{d}z$.\\
\textbf{Trainable parameters:} $\{\mW_k^Q,\mW_k^K,\cem{\mD_k=\diag(\vd_k)}\}_{k=1}^K$, $\mW,\mV$, \cem{$\{\mP_k\}_{k=1}^K,\mP_{\mathrm{mlp}}$}.
}

\vspace{1pt}
\rule{\linewidth}{0.05pt}
\noindent
\makebox[\linewidth+0.cm][l]{%
\begin{minipage}[t]{0.315\linewidth}
\centering \textbf{Main Block} \\[4pt]
\raggedright
\For{$i = 1:J$}{
  $\vh_{1:i} \gets \rms(\vh_{1:i})$ \\
  \For{$k = 1:K$}{
    $\vk_{1:i}^k \gets \mW_k^K \vh_{1:i}$ \\
    $\vv_{1:i}^k \gets \cem{\mW_k^K}\vh_{1:i}$ \\
  }
  $\vh_i \gets \operatorname{MHA}(\vh_{1:i}^{1:K},\,\vk_{1:i}^{1:K},\,\vv_{1:i}^{1:K})$ \\
  }
  \For{$i = 1:J$}{
  $\vh_{i} \gets \rms(\vh_{i})$ \\
  $\vh'_i \gets \operatorname{MLP}(\vh_i)$
}
\Return $\vh'_{1:J}$
\end{minipage}
\begin{minipage}[t]{0.395\linewidth}
\centering \textbf{Subroutine: MHA} \\[4pt]
\raggedright
\textbf{MHA}$(\vh_{1:i},\vk_{1:i}^k,\vv_{1:i}^k)$: \\
$\vx_i \gets \vh_i$ \\
\For{\cem{$t=0:T_\epsilon-1$}}{
  $\vu_i \gets \rms(\vx_i)$ \\
  \For{$k = 1:K$}{
    $\vq_i^{k} \gets \mW_k^Q \vu_i$ \\
    $\va_{ijk}\!\gets\!D_h^{\nicefrac{-1}{2}}\!
      \big(\vk_j^k\!{}^\top\!\vq_i^{k} 
      \cem{+ \vh_j^\top \mD_k \vu_i}\big)$ \\
    $\mathbf{o}_i^{k}\!\gets\!\displaystyle\sum_{j=1:i}
     \!\operatorname{softmax}_j(\{\va_{ijk}\}_{j=1}^{i})\,\vv_j^k$\\
  }
  $\vx_i \gets \vx_i + \eta_\epsilon 
      \sum_k \cem{\mP_k\,\mW_k^{Q\top}} \mathbf{o}_i^{k}$
}
\Return $\vx_i$
\end{minipage}
\hfill
\begin{minipage}[t]{0.285\linewidth}
\centering \textbf{Subroutine: MLP} \\[4pt]
\raggedright
\textbf{MLP}$(\vh_i)$: \\
$\vx_i \gets \vh_i$ \\
$\cem{\vgamma = \mW \vh_i}$\\
\For{\cem{$t=0:T_\xi-1$}}{
  $\vu_i \gets \rms(\vx_i)$ \\
  $\vg_i\gets\cem{\mV^{\!\top}}\!\!
    \big(\cem{\vgamma}\circ\phi'(\mV \vu_i\!)\!\big)$ \\
  $\vx_i \gets \vx_i + \eta_\xi \cem{\mP_{\operatorname{mlp}}}\,\vg_i$
}
\Return $\vx_i$
\end{minipage}
}
\caption{Transformer Block as Energy Updates (\cem{Orange parts} highlight \gls{cem} specifics)} \label{algo:cem_block}
\end{algorithm*}

A subtlety arises when incorporating positional encodings: rotary embeddings (RoPE) in particular complicate the energy-gradient view by making the projection weights depend on both query and key indices. To avoid this overhead, we instead adopt relative-position biases such as Alibi, as discussed in \Cref{appsec:positional-encoding}.

\section{Related Work} \label{sec:related_work}

\paragraph{Energy based sequence models.} 
\glspl{ebm} assign low energies to preferred configurations \citep{hopfield1982neural,lecun2006tutorial}. 
Modern extensions to Hopfield networks \citep{krotov2016dense,ramsauer2021hopfield} with continuous patterns and log-sum-exp energy demonstrate how attention-like updates can arise from their updating iterations. Subsequent work on sequence-level \glspl{ebm} extends these ideas to language modeling and text generation \citep{du2021improved,qin2022cold,liu2022bolt}. More recently, \citet{hoover2023energy,gladstone2025energy} explore using energy minimization as building blocks for large language models, providing a new paradigm for scaling learning and thinking capacity.
In contrast, our work does not treat energy minimization procedures as layers. Instead, we show that existing Transformer layers, including both \gls{mha} and \gls{mlp}, with weight-sharing, can already be reframed as energy-based updates. This perspective enables principled layer redesigns and extensions, and leads to improved parameter efficiency in Transformers.

\paragraph{Optimization and probabilistic-inference views of Transformers.}
Several works have connected Transformers with optimization or probabilistic
inference. \citet{yang2022transformers} view Transformers as unfolded
optimization procedures; \citet{wu2023probabilistic} derive attention-like
updates from mean-field inference; and \citet{ravuri2025transformers} interprets 
each Transformer block as performing gradient descent on a
variational lower bound of the probabilistic Laplacian Eigenmaps model. 
Closest to our work, \citet{dehmamy2025nrgpt} derive attention and feed-forward
from gradients of an unconditional
energy. In comparisons, \gls{cem} studies causal, conditional
energy updates, with particular emphasis on
the parameterizations they induce and the layer extensions they suggest.

\paragraph{Alternative transformer blocks.}  
Transformer models have largely converged on Llama-style backbones with multi-head attention \citep{vaswani2017attention} and gated \glspl{mlp} \citep{shazeer2020glu}. Many efficiency-oriented variants reduce the cost of attention via multi-query, group-query, or multi-head latent attention \citep{shazeer2019fast,ainslie2023gqa,liu2024deepseek}, while others target the feedforward block \citep{liu2021pay,shazeer2020glu,so2021primer}. Sparsely activated mixture-of-experts (MoE) layers scale capacity \citep{shazeer2017moe,fedus2022switch}, and other work simplifies skip connections, projections, or normalization with little performance loss \citep{he2024simplifying,he2023deep}. State-space models (SSMs) \citep{gu2021efficiently,gu2023mamba,yang2024parallelizing} have also shown strong results, with connections to attention noted by \citet{daotransformers}, though hybrid designs remain necessary for state-of-the-art performance \citep{team2025kimi}.

\begin{figure*}[ht!]
\centering
\begin{subfigure}[t]{0.42\textwidth}
    \centering
    \includegraphics[width=\linewidth]{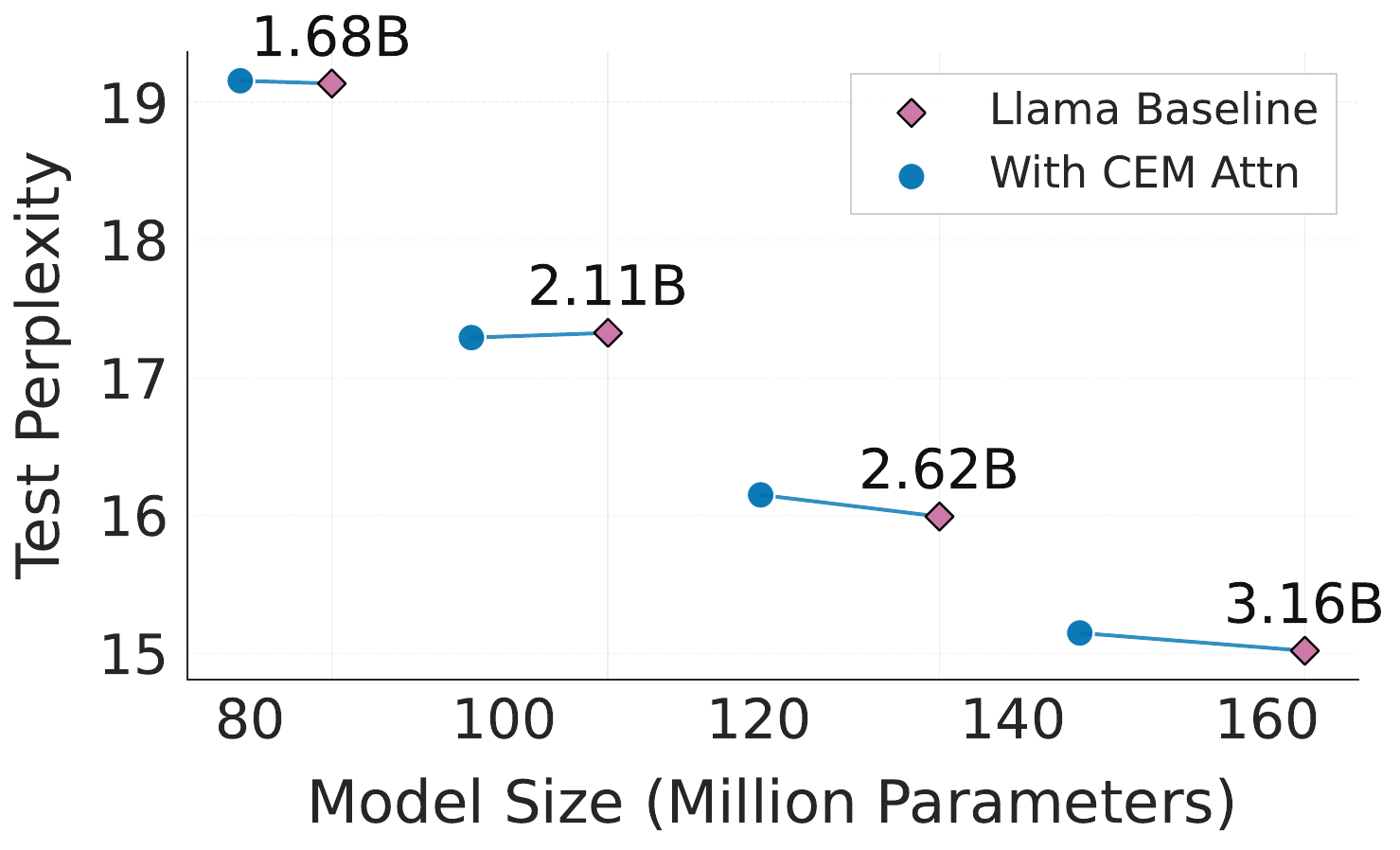}
    \caption{}\label{fig:cem_attn}
\end{subfigure}%
\begin{subfigure}[t]{0.42\textwidth}
    \centering
    \includegraphics[width=\linewidth]{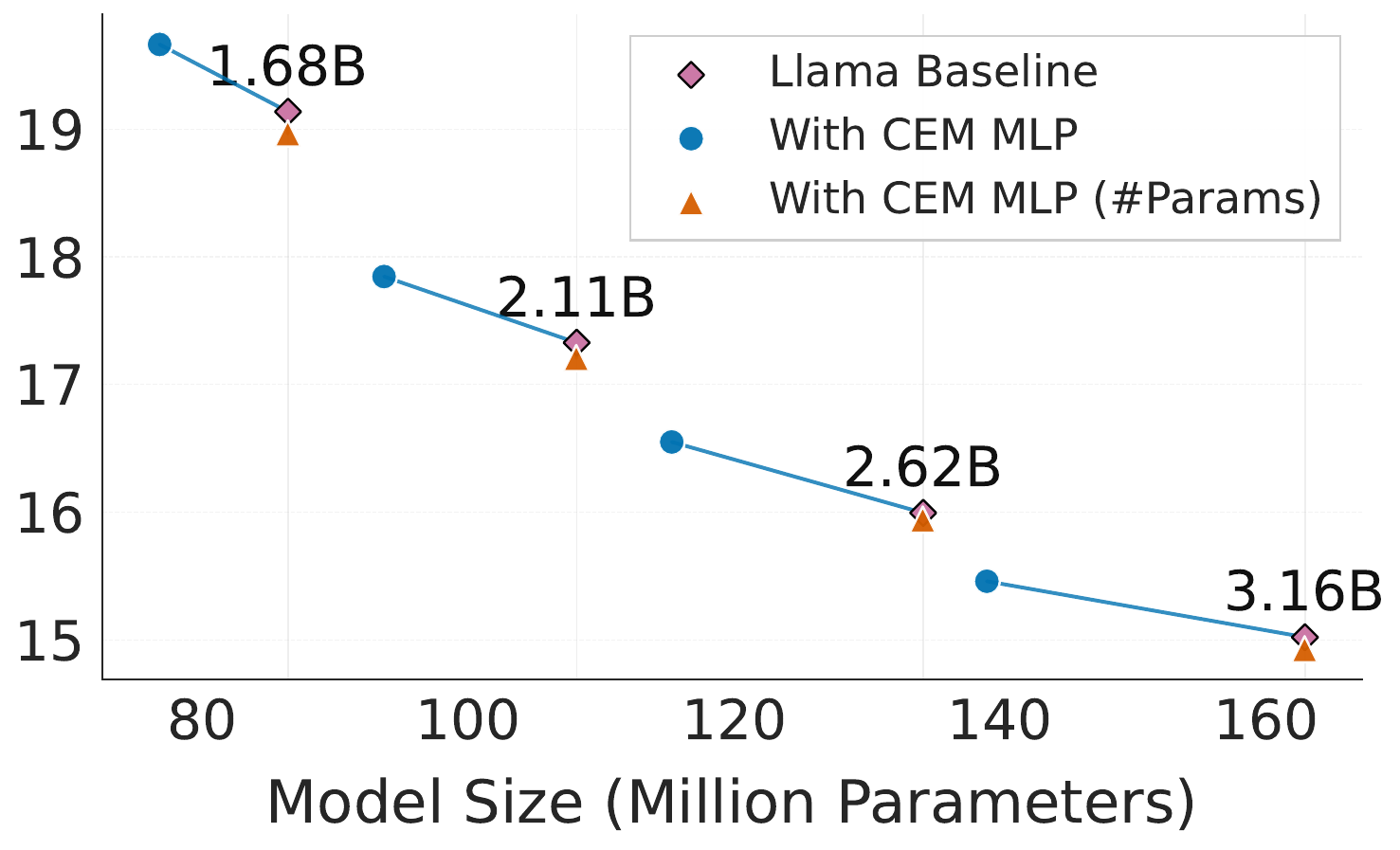}
    \caption{}\label{fig:cem_mlp}
\end{subfigure}
\hfill
\begin{subfigure}[t]{0.42\textwidth}
    \centering
    \includegraphics[width=\linewidth]{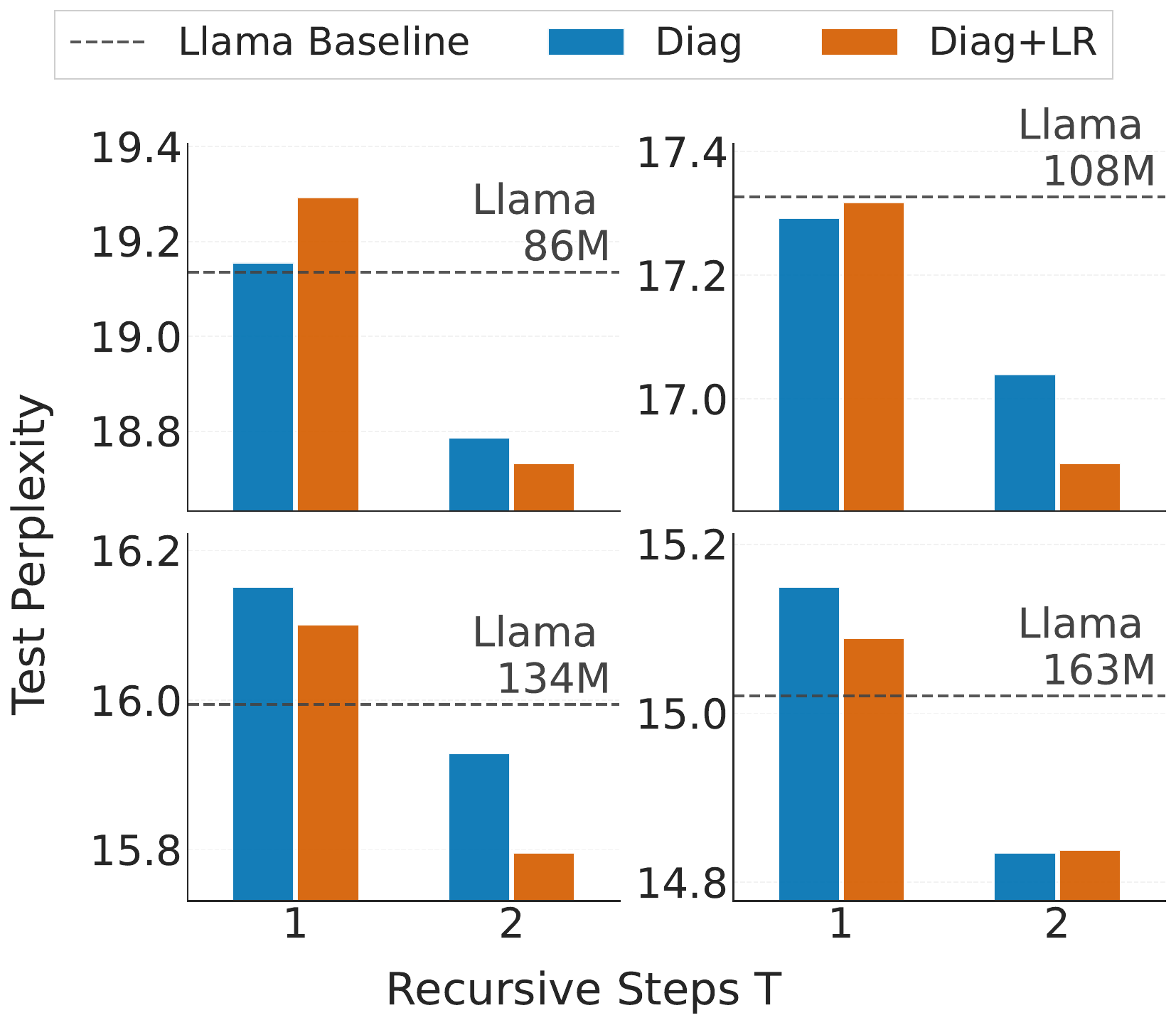}
    \caption{}\label{fig:cem_attn_T_bars}
\end{subfigure}%
\begin{subfigure}[t]{0.42\textwidth}
    \centering
    \includegraphics[width=\linewidth]{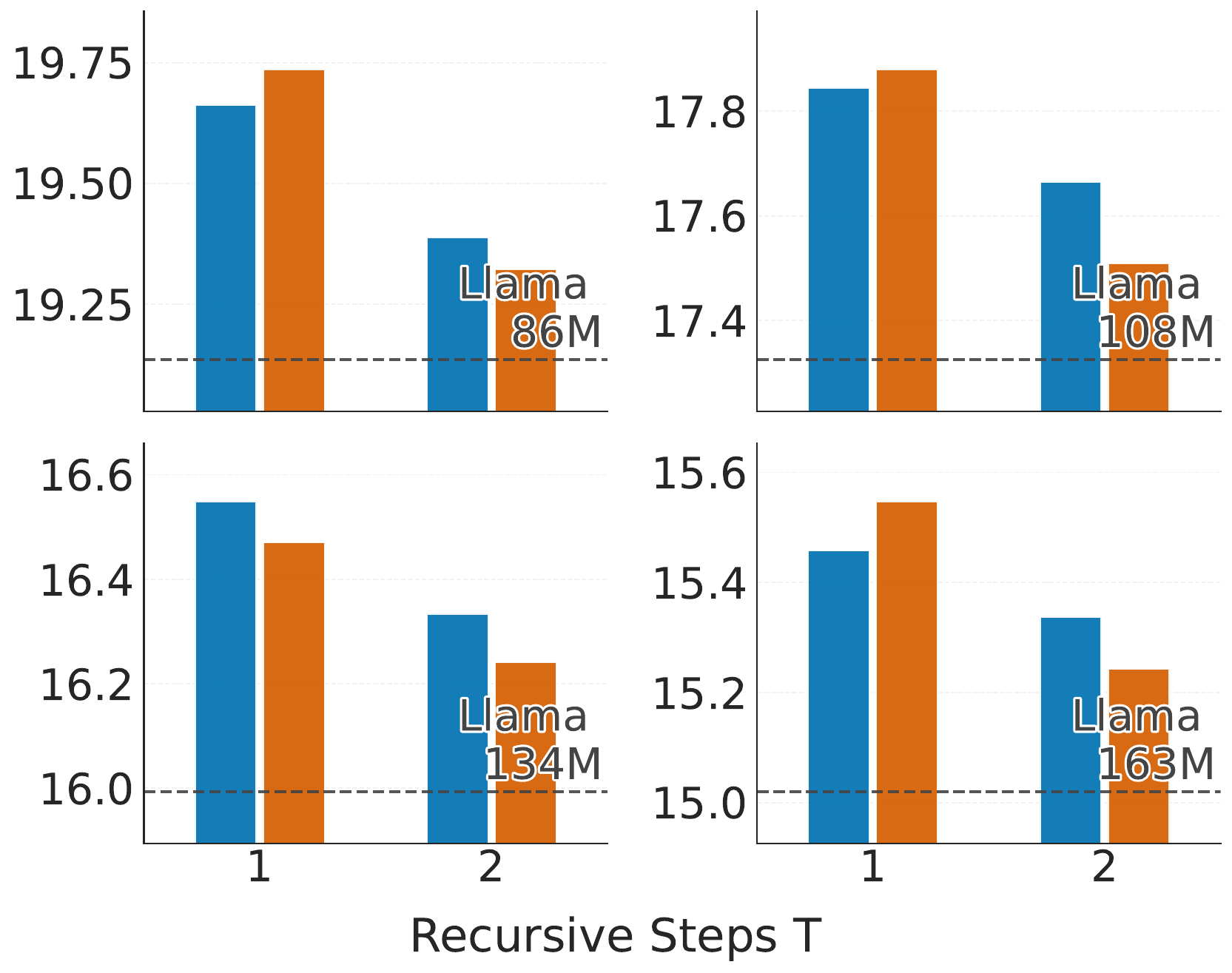}
    \caption{}\label{fig:cem_mlp_T_bars}
\end{subfigure}
\caption{(a) Llama Transformer with attention replaced by \gls{cem} attention ($T=1$). 
\gls{cem} variants (blue dots) are linked to their Llama baselines (pink diamonds) with matching dimensions but fewer parameters, trained on the same number of tokens (shown above markers).  
(b) Llama Transformer with gated \glspl{mlp} replaced by \gls{cem} \glspl{mlp} ($T=1$). Orange triangles additionally show parameter-matched variants obtained by increasing the hidden dimension.  
(c) Effects of recursion steps (x-axis) and preconditioners (colors) for \gls{cem} attention, with all models dimension-matched to Llama baselines (dashed lines).  
(d) Effects of recursion steps and preconditioners for \gls{cem} \glspl{mlp}.}
\label{fig:cem_layers}
\end{figure*}

\section{Experiments} \label{sec:experiments}
Our experiments address four questions: (i) Do the weight-sharing schemes induced by \gls{cem} lead to significant performance degradation?
(ii) do within-layer recursion and lightweight preconditioners improve performance; 
(iii) can a Transformer composed entirely of \gls{cem} layers be trained end-to-end stably; and 
(iv) how do design choices such as KQ diagonal terms and recursion affect performance. 
All models are trained on SlimPajama for the compute-optimal number of tokens of the corresponding Llama baselines \citep{hoffmann2022training}, and we report test perplexity as the main evaluation metric. Experimental details can be found in \Cref{appsec:experimental_details}.

\begin{figure*}[ht]
\centering
\begin{subfigure}[t]{0.42\textwidth}
    \centering
    \includegraphics[width=\linewidth]{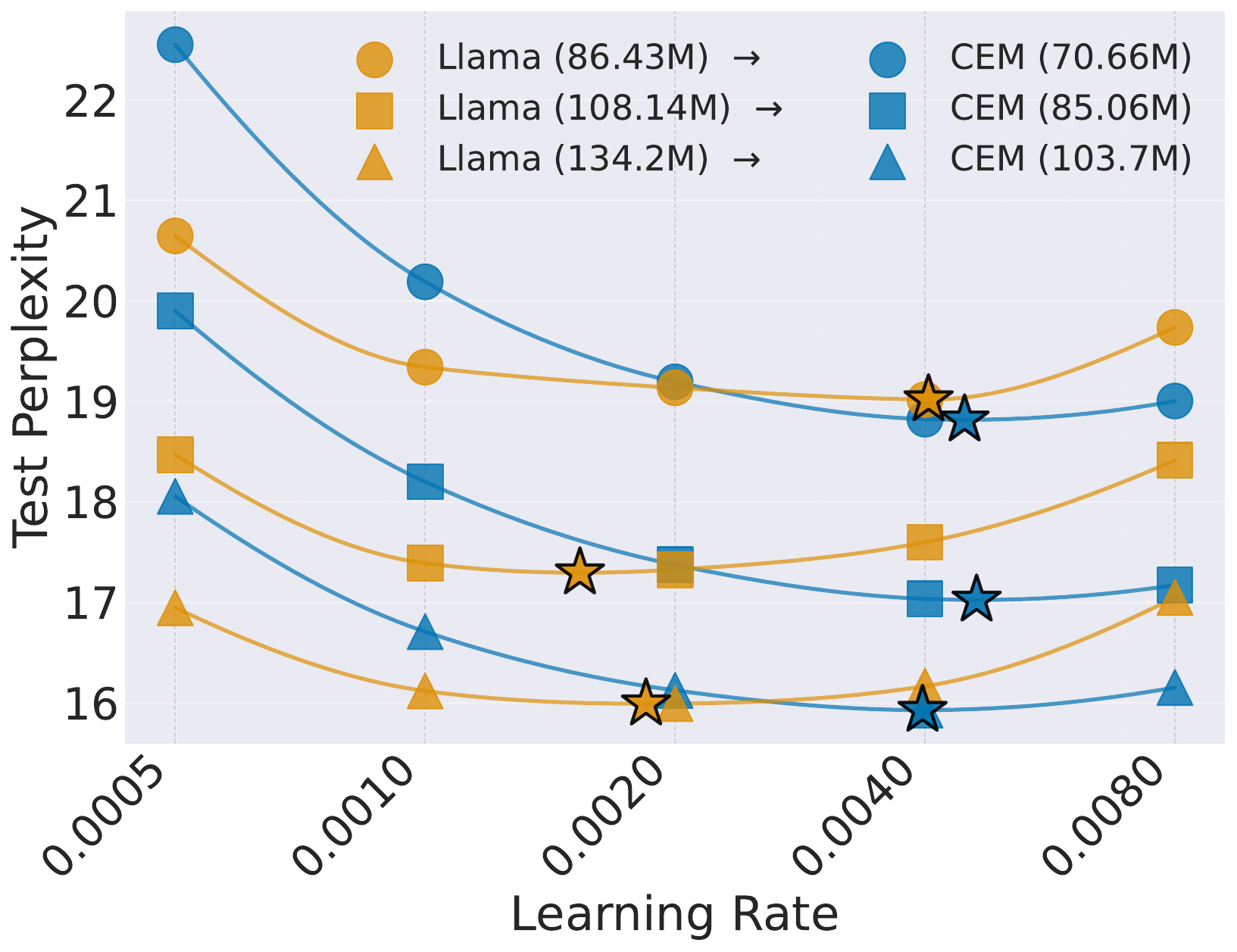}
    \caption{} \label{fig:akima}
\end{subfigure}%
\begin{subfigure}[t]{0.42\textwidth}
    \centering
    \includegraphics[width=\linewidth]{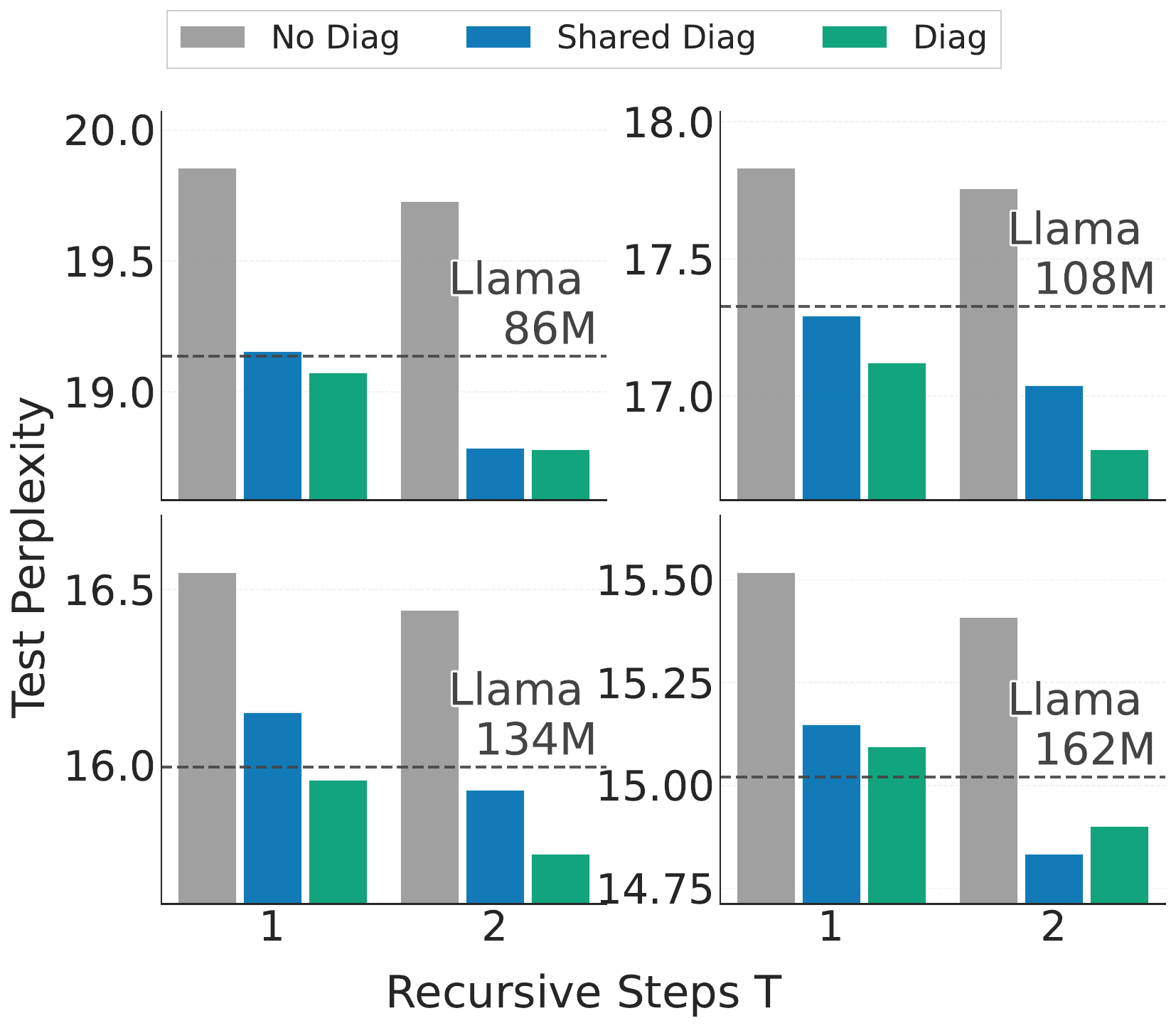}
    \caption{} \label{fig:kq_diag}
\end{subfigure}%
\hfill
\begin{subfigure}[t]{0.95\textwidth}
    \centering
    \includegraphics[width=\linewidth]{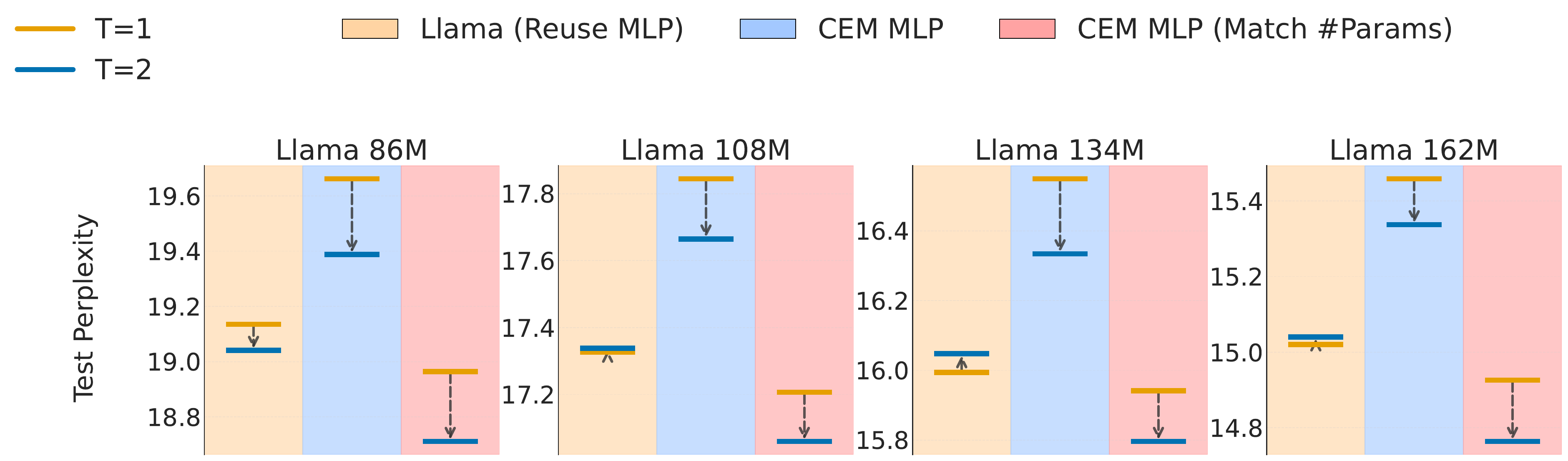}
    \caption{} \label{fig:recursive_mlp}
\end{subfigure}
\caption{(a) Optimal learning rate estimated via Akima interpolation. 
Orange denotes baseline Llama models and blue denotes \gls{cem} models ($T=2$) with both MHA and \gls{mlp} replaced. 
Marker shapes indicate model size; stars mark interpolated optima from five data points. 
Matched Llama and \gls{cem} models (with roughly half the MHA and two-third the \gls{mlp} parameters) are trained with the same token budget (Chinchilla-optimal for Llama). 
For smaller models, parameter reduction is less pronounced due to embedding and head parameters. 
(b) KQ diagonal strategies in \gls{cem} attention: no diagonal in $\mA_k$, a shared diagonal across heads, and per-head diagonals. 
All \gls{cem} models match the dimensionality of the Llama baselines (dashed line). 
(c) Within-layer recursion vs.\ plain layer reuse in \glspl{mlp}. We compare the performance gains of increasing recursion from $T=1$ (orange) to $T=2$ (blue), under three settings: Plain layer reuse (light orange area), dimension-matched \gls{cem} \gls{mlp} (blue area) and parameter-matched \glspl{mlp} (pink area). An equivalent figure comparing within-layer recursion vs.\ layer reuse for \gls{mha} can be found in \Cref{fig:mha_recursion}.}
\label{fig:ablation}
\end{figure*}

\subsection{Replace Transformer layers with single-step CEM layers}
To evaluate the effectiveness of \gls{cem} layers, we train Transformer models with \gls{cem} components in either the \gls{mlp} or attention blocks, and compare against Llama baselines. We focus on the weight-tying formulation (see Equations~\ref{eq:attn_weights_sharing} and \ref{eq:mlp_weights_sharing}), but without recursions or preconditioners here.

\Cref{fig:cem_attn} compares \gls{cem} attention with standard Llama MHAs, while \Cref{fig:cem_mlp} compares \gls{cem} \glspl{mlp} with Llama-style gated \glspl{mlp}. Blue dots denote dimension-matched \gls{cem} models, where \gls{cem} attention uses about half the parameters and \gls{cem} \glspl{mlp} about two-thirds of their Llama counterparts. Some degradation is expected, but the goal is to assess how closely \gls{cem} models approach baseline performance with fewer parameters. For \gls{cem} \glspl{mlp}, we also report results with increased intermediate dimension to restore the baseline parameter count (orange triangles).

Replacing attention with the \gls{cem} variant has only a small effect on test perplexity despite halving the parameter count, with no natural parameter-matching scheme available since the model dimension must remain fixed for controlled comparison. For \gls{cem}-\glspl{mlp}, perplexity is higher due to parameter sharing, but increasing the hidden dimension to match parameter count yields consistent, albeit modest, improvements in perplexity — though at the cost of additional FLOPs. Unless otherwise noted, we adopt the optimal Llama hyperparameters from grid search (see \Cref{appsec:experimental_details}) to make consistent comparisons and avoid tuning each \gls{cem} configuration individually, even though these settings may be suboptimal for \gls{cem} models (see \Cref{fig:akima}).
These results indicate that single-step \gls{cem} layers define constrained,
parameter-efficient variants of standard Transformer components that can retain
competitive perplexity in this controlled setting. The results are strongest
for \gls{cem} attention, suggesting that the corresponding weight-sharing
parameterization merits further investigation.

\subsection{Recursive steps and preconditioners in CEM layers}
We test whether within-layer recursion and lightweight preconditioners improve performance. Transformer variants are trained on SlimPajama to the compute-optimal token budget of their Llama baselines and evaluated by test perplexity. \Cref{fig:cem_attn_T_bars} and \Cref{fig:cem_mlp_T_bars} replace the standard Llama \glspl{mha} and \glspl{mlp} with their \gls{cem} counterparts, respectively. We compare diagonal and diagonal-plus-low-rank preconditioners for $T=1$ and $T=2$. All models are dimension-matched to their Llama baselines, so \gls{cem} components use fewer parameters, and preconditioners add negligible overhead.

As shown in \Cref{fig:cem_layers}, the two ingredients contribute unequally.
Increasing recursion from $T=1$ to $T=2$ consistently improves \gls{cem}
layers, and for attention, recursive \gls{cem} variants can outperform the
Llama baseline while using fewer parameters. For \glspl{mlp}, \gls{cem}
variants remain below the baseline, but the gap narrows at $T=2$. Learned
preconditioners, by contrast, contribute only marginally: they cannot 
provide consistent gains especially for $T=1$, suggesting that
the recursive structure, rather than the preconditioner, drives most of the
improvement.

We did not observe consistent gains for $T \geq 3$ under our standard training setup, likely due to the difficulty of optimizing through additional unrolled iterations. However, controlled experiments up to $T=8$ on a synthetic problem (\Cref{tab:appendix_full_s1}) show that deeper recursion can yield further improvements. Thus, scaling beyond $T=2$ in full Transformer settings remains promising but will likely require advances in optimization for unrolled architectures, which we leave to future work.

\subsection{Training Transformers with CEM-derived layers}
We next test whether a Transformer composed entirely of CEM-derived attention and
MLP layers can be trained end-to-end. We use $T=2$ with diagonal-plus-low-rank
preconditioners for both components, which keeps the constrained \gls{cem}
parameterization: about half the attention parameters and two-thirds the MLP
parameters of the standard counterparts. Due to memory constraints, we omit the
largest model with diagonal-plus-low-rank preconditioners.

We sweep five learning rates from $0.0005$ to $0.008$ and use Akima interpolation~\citep{Akima1970ANM} to estimate the optimum. \Cref{fig:akima} reports the interpolated test perplexity. We additionally report a $256$M-parameter setting in \Cref{fig:akima_plot_256m} of the Appendix, where the same observation holds.

End-to-end \gls{cem} models train stably and achieve perplexity comparable to
the corresponding Llama baselines while using fewer parameters. The interpolated
optima suggest that \gls{cem} models may prefer higher learning rates, though
this trend requires further study.

\subsection{Ablation study}
We first study the role of diagonal terms in inter-token distances (\Cref{fig:kq_diag}) in attention, comparing three settings: no diagonal, a shared diagonal across heads, and per-head diagonals. All other components are fixed (Llama \glspl{mlp}, \gls{cem} attention with one recursion step, and a simple diagonal preconditioner). Including a diagonal term proves essential for good performance, and our shared-diagonal strategy provides performance close to per-head diagonals while reducing parameters and compute, making it a more efficient alternative.  

Second, we test whether within-layer recursion is necessary or if naive layer reuse suffices (\Cref{fig:recursive_mlp}). Simply reapplying the same residual block yields little or no perplexity gain. Note that this reuse differs from recursive Transformer, where entire blocks (attention and MLP) are reused. In contrast, CEM-based within-layer recursion produces consistent improvements in both dimension- and parameter-matched settings. A similar trend holds for attention (\Cref{fig:mha_recursion}).

\section{Discussion and Conclusion}

\subsection{Limitations} \label{subsec:limitations}
This work studies Transformer layer parameterization through the lens of energy updates, but developing scalable and performant new architectures will require further investigation. Our experiments therefore provide a controlled evaluation at the hundred-million-parameter scale, using test perplexity as a standard proxy for language-model quality. Larger-scale scaling studies and downstream evaluations, which become most meaningful at larger model sizes, are natural next steps for assessing how the benefits of \gls{cem}-derived layers transfer to larger models and practical tasks. In addition, improving the efficiency of these layers will likely require custom kernels and systems-level optimization. Our implementation focuses on validating the proposed parameterizations rather than optimizing runtime performance. Because \gls{cem}-derived layers introduce structured weight sharing and recursive updates, custom kernels, fused operations, and hardware-aware implementations will be needed to improve their practical runtime behavior.

\subsection{Conclusion and future directions} \label{subsec:conclusion_future_directions}

We introduced \gls{cem}, a framework that recasts Transformer layers as causal energy minimization. From this view, weight-tied attention and gated \glspl{mlp} emerge as energy-gradient updates, motivating optimization-inspired extensions such as diagonal-plus-low-rank parameterization, lightweight preconditioners, and recursive updates. These \gls{cem}-derived layers approach or match Transformer baselines in moderate-scale language modeling, with recursion and diagonal-plus-low-rank parameterizations yielding the most consistent gains and preconditioners providing more marginal improvements. Overall, \gls{cem} offers a new lens for rethinking Transformer parameterizations.

We believe the following directions are particularly promising:
\begin{enumerate*}[label=(\arabic*)]
\item \textbf{Custom kernels}, where FlashAttention-style IO-aware implementations \citep{dao2022flashattention} could fuse tied projections, diagonal-plus-low-rank terms, and recursive updates to improve throughput; 
\item \textbf{Test-time scaling}, where \gls{cem}-style within-layer recursion may provide an additional axis for scaling test-time compute \citep{snell2024scaling,muennighoff2025s1simpletesttimescaling} and for supporting latent reasoning \citep{hao2024coconut,zhang2024uncovering,tan2025think}, especially when combined with blockwise recursion as in looped or recursive Transformers \citep{yang2023looped,bae2024relaxed,dehghani2018universal}; and
\item \textbf{Architecture--hardware co-design}, where alternative optimization procedures for the same underlying energy could yield layer parameterizations that are jointly designed with new hardware specifics and kernel implementation.
\end{enumerate*}

\begin{ack}
We would like to thank Riccardo Grazzi, Elon Portugaly, Babak Rahmani, Jannes Gladrow, Taketomo Isazawa, and many others at Microsoft Research Cambridge for their valuable discussions and early feedback on this work. We also thank the anonymous reviewers for their constructive suggestions, which helped improve the clarity and presentation of the paper.
\end{ack}


\bibliography{cem}
\bibliographystyle{plainnat}

\newpage
\appendix

\section{Background} \label{appsec:background}

\subsection{Equivalence between concatenation and summation in attention} \label{subsec:concatenation_summation}

In \Cref{subsec:attention-as-gradient}, we write the multi-head attention update in the form
\[
    \mha(\vh_{1:i})
    = \sum_{k=1}^K \mW_k^{O\top}
        \Bigg( \sum_{j=1}^i 
            \softmax_j\!\Big( \big\{ \tfrac{1}{\sqrt{D_h}} (\vk_{j'}^k)^\top \vq_i^k \big\}_{j'=1}^i \Big)\,
            \vv_j^k
        \Bigg),
\]
where each head \(k\) contributes an output vector that is multiplied by a head-specific block 
\(\mW_k^O \in \mathbb{R}^{D_r \times D_h}\).

This notation differs slightly from the conventional implementation of multi-head attention, where per-head outputs are concatenated and processed by a single output projection. To make the equivalence explicit, let
\[
    O_k \;\coloneqq\; \sum_{j=1}^i 
        \softmax_j\!\Big( \big\{ \tfrac{1}{\sqrt{D_h}} (\vk_{j'}^k)^\top \vq_i^k \big\}_{j'=1}^i \Big)\,
        \vv_j^k
    \in \mathbb{R}^{D_r}
\]
denote the output of head \(k\). The standard formulation concatenates these outputs,
\[
    O = [\,O_1;\,O_2;\,\dots;\,O_K\,] \in \mathbb{R}^{K D_r},
\]
where $;$ denotes vertical stacking of matrices,
and applies a single output projection \(\mW^{O} \in \mathbb{R}^{K D_r \times D_h}\).

If we partition \(\mW^O\) into head-aligned blocks,
\[
    \mW^O = [\, \mW_1^{O};\; \mW_2^{O};\; \dots;\; \mW_K^{O} \,],
    \qquad
    \mW_k^O \in \mathbb{R}^{D_r \times D_h},
\]
then multiplying out gives
\[
    \mW^{O\top} O 
    = [\,\mW_1^{O\top},\dots,\mW_K^{O\top}\,]
      [\,O_1;\,O_2;\,\dots;\,O_K\,]
    = \sum_{k=1}^K \mW_k^{O\top}\, O_k.
\]

Thus, the conventional \emph{concatenation followed by a single projection} is algebraically equivalent to the \emph{sum over head-specific projections} used in our presentation. We adopt the latter form for notational clarity in the CEM formulation.

\subsection{Diagonal-plus-low-rank parameterization} \label{diag-plus-low-rank-parameterization}

We use the diagonal-plus-low-rank (D+LR) parameterization for both the
attention-score computation (in \Cref{eq:diag_plus_low_rank_kv}) and the preconditioners (in \Cref{eq:diag_plus_low_rank_preconditioners}. Here we provide a brief
background on this parameterization.

The basic form of a D+LR matrix is often written as
\[
W = \mathrm{diag}(d) + UV^\top,\qquad U,V\in\mathbb{R}^{d\times r},\quad r\ll d.
\]
Compared with a pure diagonal parameterization, which cannot express
cross-feature interactions, and a pure low-rank parameterization, which
only captures interactions within a rank-$r$ subspace, the D+LR form models
both aspects. 
The memory footprint and computational cost are still much smaller compared to the full matrix: applying
$W$ requires one diagonal pass and two thin matrix multiplications, with
complexity $O(d) + O(dr)$, far smaller than the $O(d^2)$ cost of a dense
matrix. 
D+LR parameterizations have already been widely used in deep learning such as ~\citep{gu2021efficiently,bonnabel2024low}.

\section{Incorporating positional encoding into CEM attention} \label{appsec:positional-encoding}

\paragraph{Positional encoding.}  
Standard Transformer architectures such as Llama employ Rotary Position Embeddings (RoPE) \citep{su2024roformer} to encode relative position information.  
Recall from \Cref{subsec:attention-as-gradient} that in our energy-based formulation, each head $k$ is parameterized by a matrix $\mA_k$:
\[
    \vbeta_{kj} = \mA_k \vh_j, \qquad \text{with } \mA_k \in \sR^{D_h \times D_h}.
\]
In the simplest case, we adopt a low-rank factorization $\mA_k = \mW_k^{Q\top}\mW_k^K$, so that queries, keys, and values arise as
\[
    \vq_i^k = \mW_k^Q \vh_i, 
    \quad \vk_j^k = \mW_k^K \vh_j, 
    \quad \vv_j^k = \mW_k^V \vh_j,
\]
under the weight-tying constraints $\mW_k^K = \mW_k^V$ and $\mW_k^Q = \mW_k^O$ (see \eqref{eq:attn_weights_sharing}).  
The interaction energy is then defined as
\[
    \epsilon(\vx_i \mid \vh_{1:i}) 
    = -\tau \sum_{k=1}^K \log\sum_{j=1}^i 
        \exp\!\Big( \tfrac{1}{\tau}\,\vbeta_{kj}^{\top}\vx_i\Big),
\]
and its gradient update recovers the standard multi-head attention form with weight sharing.  

When incorporating RoPE, however, $\mA_k$ must depend explicitly on both indices $i$ and $j$ through rotation matrices $\mR(i)$ and $\mR(j)$:
\[
    \mA_k \;=\; \mW_k^{Q\top}\, \mR(i)^\top \mR(j)\, \mW_k^K .
\]
This makes $\vbeta_{kj}$ dependent on the query index $i$ as well as $j$, which substantially increases memory costs: the value projection effectively becomes query-dependent.

\paragraph{Alibi positional encodings.}  
To mitigate this overhead, we instead adopt \newterm{Alibi positional encodings} \citep{press2022train}, which introduce a head-specific bias
\[
    b_{ijk} = -m_k |i-j|
\]
directly into the attention scores before the softmax.  
Concretely, in the unbiased case the score is
\[
    s_{ijk} = \tfrac{1}{\tau}\, \vbeta_{kj}^\top \vx_i,
\]
so with Alibi it becomes
\[
    s_{ijk} \;=\; \tfrac{1}{\tau}\,\vbeta_{kj}^\top \vx_i \;+\; b_{ijk},
\]
and the normalized weights are
\[
    \alpha_{ij}^k = \softmax_j \!\Big(\{ s_{ij'k} \}_{j'=1}^i \Big).
\]
The slopes $m_k$ are typically chosen as a geometric sequence, e.g.\ $m_k = 2^{-k}$.  
This adds negligible overhead compared to RoPE while still encoding relative bias.  
In practice, we further include a learnable bias distinguishing self- vs.\ cross-token attention:
\[
    b_{ijk} = -m_k |i-j| + b_{i=j} + b_{i\ne j}.
\]

\paragraph{Interaction energy with bias.}  
In the energy formulation, this simply shifts the logits inside the log-sum-exp:
\[
    \epsilon(\vx_i \mid \vh_{1:i}) 
    = -\tau \sum_{k=1}^K \log \sum_{j=1}^i 
        \exp\!\Big( \tfrac{1}{\tau}\,\vbeta_{kj}^\top \vx_i + b_{ijk}\Big).
\]
The corresponding gradient update is
\[
    \nabla_{\vx_i}\, \epsilon(\vx_i \mid \vh_{1:i})
    = -\sum_{k=1}^{K} \sum_{j=1}^i 
    \softmax_j \!\Big( \tfrac{1}{\tau}\,\vbeta_{kj}^\top \vx_i + b_{ijk}\Big)\, \vbeta_{kj},
\]
so $b_{ijk}$ modifies the logits before normalization but leaves the overall gradient structure unchanged.

\section{Relation to Hopfield networks} \label{appsec:hopfield}
Hopfield networks are classical models of associative memory, where stored patterns correspond to attractors of an energy landscape, and the dynamics converge to the attractor most consistent with the initial state.
This viewpoint aligns with our interpretation of Transformer layers as energy-minimizing updates: both attention and \gls{mlp} sublayers can be seen as iterative steps that decrease a suitably defined energy function.
We next detail these connections.

\paragraph{Interaction energy.}  
Classical Hopfield networks \citep{hopfield1982neural} store a finite set of patterns $\{\vh_j\}$ in an energy function of the form
\[
    \epsilon(\vx) = - \tfrac{1}{2}\sum_j (\vh_j^\top \vx)^2,
\]
More recent extensions reinterpret Hopfield networks as continuous attractor models, greatly expanding their representational capacity.  
For instance, dense associative memories \citep{krotov2016dense} and modern Hopfield networks \citep{ramsauer2021hopfield} introduce an energy of the log-sum-exp form,  
\[
    \epsilon(\vx) = - \tau\log \sum_j \exp\!\Big(\tfrac{1}{\tau}\vh_j^\top \vx\Big),
\]
which is convex in $\vx$ and whose fixed-point updates under the concave-convex procedure (CCCP) \citep{Yuille2001TheCP} yield
\[
    \vx' = \sum_j \softmax_j\!\Big(\tfrac{1}{\tau}\vh_j^\top \vx\Big)\,\vh_j,
\]
exactly the update rule underlying the attention mechanism.  
This connection underlies the interpretation of attention as a form of fast Hopfield retrieval.

\paragraph{Our perspective.} We depart from the setup of modern Hopfield networks in three important ways. First, instead of computing fixed points via iterative CCCP updates \citep{Yuille2001TheCP}, we interpret each Transformer sublayer as performing a \emph{single gradient step} on an energy function. Second, in our formulation the query and key projection matrices are embedded directly in the energy, which causes them to reappear as the output--value projections in the gradient update---naturally yielding the tied $\mW_Q, \mW_K$ and $\mW_O, \mW_V$ structure of attention. Finally, while \citet{ramsauer2021hopfield} introduce novel Hopfield layers and evaluate them on associative-memory benchmarks, our framework treats standard Transformer layers themselves as energy-based updates, and we demonstrate that this perspective leads to principled extensions and improvements for text modeling tasks.

\paragraph{Element-wise energy.}  
The element-wise energy leading to gated \glspl{mlp} has a less direct connection.  
Optimization via CCCP is possible only when using a convex form. We briefly experimented with models using energies of the form  
\[
    \xi(\vx_i \mid \vh_i) 
    = -|\vgamma_i|^\top \,\phi\big(\diag(\operatorname{sign}(\vgamma_i))\,\mV \vx_i\big), 
    \qquad \vgamma_i = \mW \vh_i,
\]
with $\phi$ a convex nonlinearity, so that the energy is convex in $\vx$. The gradient of this energy form is
\[
 - \mV^\top \left(\vgamma_i \circ \phi'(\diag(\sign(\vgamma_i))\mV\vx_i)\right)
\]We used a straight-through estimator to deal with the sign nonlinearity. We found that these models successfully trained, but with worse performance than ignoring the sign.  
Unlike the interaction energy, the link to memory association here is unclear, as are the corresponding convergence guarantees and capacity limits.

\section{Experimental details}
\label{appsec:experimental_details}

\subsection{Model architectures}

We evaluate \gls{cem}-based architectures across multiple model scales ranging from 86M to 162M parameters. All models follow the Llama architecture as baseline with modifications for \gls{cem} components. \Cref{tab:model_configs} summarizes the architectural details for each model size.

\begin{table}[h]
\centering
\caption{Model architecture configurations for different parameter counts. All models use a vocabulary size of 32,000 tokens.}
\label{tab:model_configs}
\begin{tabular}{lccccc}
\toprule
\textbf{Configuration} & \textbf{86M} & \textbf{108M} & \textbf{134M} & \textbf{162M}  \\
\midrule
Model dimension ($d_h$) & 672 & 672 & 768 & 864  \\
Number of layers & 8 & 12 & 12 & 12  \\
Number of heads & 8 & 12 & 12 & 12  \\
\gls{mlp} intermediate dimension & 1792 & 1792 & 2048 & 2304  \\
Context length & 2048 & 2048 & 2048 & 2048  \\
\midrule
\end{tabular}
\end{table}

\subsection{Dataset and preprocessing}
\label{appsec:data_preprocessing}

\paragraph{Dataset.}  
We use a subset of SlimPajama-627B \citep{cerebras2023slimpajama}, a cleaned and deduplicated variant of RedPajama comprising approximately 627 billion tokens drawn from CommonCrawl, C4, GitHub, books, arXiv, Wikipedia, and StackExchange. The dataset is accessed via \texttt{gmongaras/SlimPajama-627B\_Reupload} on Hugging Face and is released under the Apache 2.0 license.

\paragraph{Tokenization.}  
We employ the \texttt{LlamaTokenizerFast} with a vocabulary size of 32,000 tokens.  

\paragraph{Data processing.}  
Documents are concatenated and split into fixed-length sequences of 2048 tokens, with no padding applied.  

\subsection{Training configuration}
Training hyperparameters are summarized in Table~\ref{tab:training_config}. We follow Chinchilla-optimal compute allocation \citep{hoffmann2022training} for determining the number of training tokens for each model size.

\begin{table}[h]
\centering
\caption{Training hyperparameters for \gls{cem} models and Llama baselines.}
\label{tab:training_config}
\begin{tabular}{lcc}
\toprule
\textbf{Hyperparameter} & \textbf{\gls{cem} models} & \textbf{Llama baseline} \\
\midrule
Optimizer & \multicolumn{2}{c}{AdamW} \\
Learning rate & \multicolumn{2}{c}{0.002} \\
$\beta_1$ & \multicolumn{2}{c}{0.9}  \\
$\beta_2$ & \multicolumn{2}{c}{0.95}  \\
$\epsilon$ & \multicolumn{2}{c}{1e-9}  \\
Weight decay & \multicolumn{2}{c}{0.1} \\
Gradient clipping & \multicolumn{2}{c}{1.0}  \\
\midrule
LR schedule & \multicolumn{2}{c}{Cosine} \\
Warmup steps & \multicolumn{2}{c}{5\% of total} \\
Final LR factor & \multicolumn{2}{c}{0.1} \\
\midrule
Batch size (per GPU) & \multicolumn{2}{c}{8}  \\
Gradient accumulation & \multicolumn{2}{c}{4}  \\
Effective batch size & \multicolumn{2}{c}{128} \\
Precision & \multicolumn{2}{c}{bf16-mixed}  \\
\bottomrule
\end{tabular}
\end{table}

\subsection{Initialisation of preconditioners}

In \Cref{subsec:enhance_layer_optimization_perspective}, we introduce a trainable diagonal-plus-low-rank preconditioner of the form
\[
    \mP = \diag\!\big(\softplus(\vd)\big) + \mU\mV^\top + \mV\mU^\top.
\]
with $\vd \in \R^{D_h}$ and $\mU,\mV \in \R^{D_h \times R}$, where $R \ll D_h$.
Following \citet{hu202lora}, we initialize $\mU$ from a normal distribution ($\sigma=0.02$) and set $\mV$ to zeros. For the diagonal term, we parameterize
\[
    \vd = \sqrt{D_h}\,\vp,
\]
where $\vp$ is initialized to $1/\sqrt{D_h}$. This ensures that $\vd$ starts at $1$, but still yielding an appropriate effective gradient step size.

To keep the preconditioners lightweight, we set $R=4$ for attention modules and $R=16$ for MLPs. In the diagonal-only case, the preconditioner reduces to  
\[
    \mP = \diag\!\big(\softplus(\vd)\big).
\]

\subsection{Compute resources} \label{appsubsecLcompute_resources}
All experiments were conducted on a cluster of $8\times$~NVIDIA A100 GPUs (40GB memory each).  
Training time per model scales with size: the smallest models ($\sim$86M parameters) require about $8\times 2$ GPU-hours, while the largest models we tested ($\sim$162M parameters) require about $8\times 18$ GPU-hours.  
End-to-end reproduction of all results in this paper would therefore require on the order of $10{,}000$ GPU-hours.

\section{Additional results} \label{appsec:additional_results}

\begin{figure}[ht!]
\begin{center}
\includegraphics[width=0.98\textwidth]{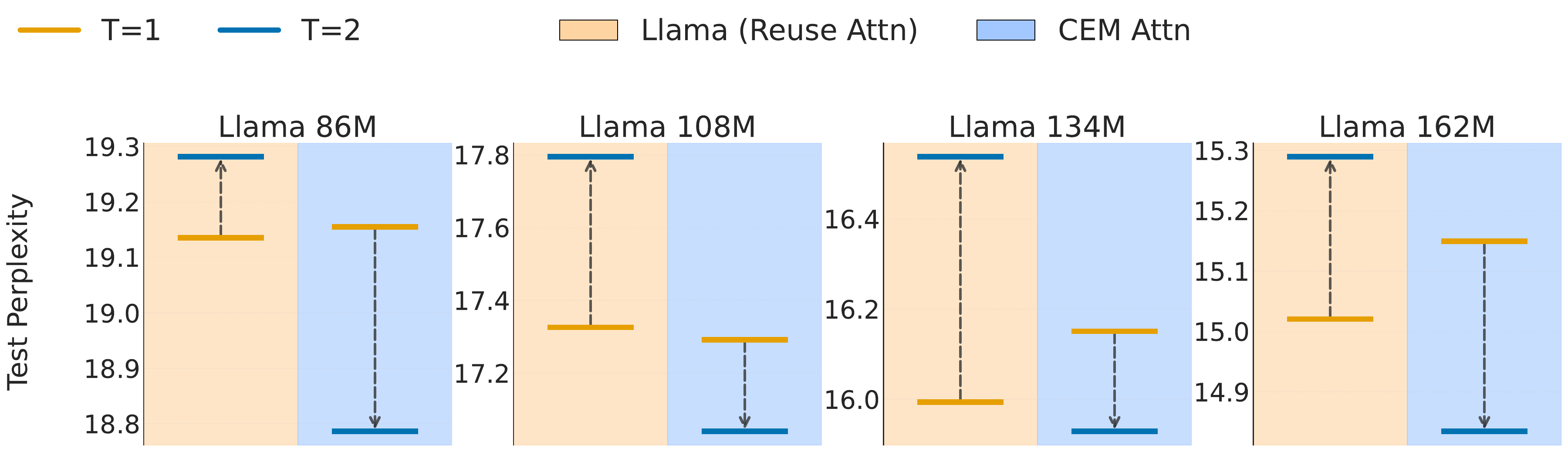}
\end{center}
\caption{Within-layer recursion vs. plain layer reuse in MHAs. We compare the performance gains of increasing recursion from $T=1$ (orange) to $T=2$ (blue), under two settings: Plain layer reuse (light orange area), dimension-matched \gls{cem} MHA (blue area).}  
\label{fig:mha_recursion}
\end{figure}

\textbf{Recursive updates in MHA}
Similar to our analysis of \gls{mlp} recursion~\Cref{fig:recursive_mlp}, we examine recursive updates in attention layers (\Cref{fig:mha_recursion}). As with \glspl{mlp}, naive reuse of the same \gls{mha} block offers no benefit and can even degrade performance in the case of \gls{mha}. In contrast, within-layer recursion in \gls{cem} attention yields clear and consistent perplexity improvements.

\textbf{Study recursion on synthetic data}
To better isolate and understand the intrinsic behaviour of the recursion, we also examine it in a controlled and computationally lightweight setting using Gaussian-process generated data and fit with our recursive CEM MLP. The results in \Cref{tab:appendix_full_s1}, illustrate that additional recursive steps generally improve performance, though the gains are not strictly monotonic.

\begin{table}[ht]
\centering
\caption{%
Model size, compute, and RMSE (mean $\pm$ std) on synthetic data sampled from Gaussian processes with 10 input dimensions. 
Here the latent state dimension is set equal to the model dimension. 
Values are averaged over repeated runs, and the lowest (best) train and test RMSE for each kernel are highlighted in \textbf{bold}. 
FLOPs are reported per forward pass. (Abbreviations: K $=$ $10^3$, M $=$ $10^6$.)}
\label{tab:appendix_full_s1}
\footnotesize
\setlength{\tabcolsep}{6pt}
\renewcommand{\arraystretch}{1.2}
\begin{tabular}{lccccc}
\toprule
\textsc{Model} & \textsc{Parameters} & \textsc{FLOPs} & \textsc{Kernel} & \textsc{Train RMSE} & \textsc{Test RMSE} \\
\midrule
Plain    & 33.98K & 6.76M  & RBF                & 0.0139 $\pm$ 0.0022 & 0.0158 $\pm$ 0.0024 \\
Gated    & 33.94K & 6.73M  & RBF                & 0.0109 $\pm$ 0.0014 & 0.0122 $\pm$ 0.0015 \\
\gls{cem}-T1   & 22.66K & 6.73M  & RBF                & 0.0102 $\pm$ 0.0010 & 0.0119 $\pm$ 0.0011 \\
\gls{cem}-T2   & 22.66K & 11.08M & RBF                & 0.0074 $\pm$ 0.0008 & 0.0092 $\pm$ 0.0008 \\
\gls{cem}-T4   & 22.66K & 19.79M & RBF                & 0.0068 $\pm$ 0.0006 & \textbf{0.0086 $\pm$ 0.0007} \\
\gls{cem}-T8   & 22.66K & 37.20M & RBF                & \textbf{0.0066 $\pm$ 0.0008} & 0.0088 $\pm$ 0.0012 \\
\midrule
Plain    & 33.98K & 6.76M  & Matern             & 0.2129 $\pm$ 0.0205 & 0.2308 $\pm$ 0.0237 \\
Gated    & 33.94K & 6.73M  & Matern             & 0.1903 $\pm$ 0.0205 & 0.2162 $\pm$ 0.0253 \\
\gls{cem}-T1   & 22.66K & 6.73M  & Matern             & 0.1836 $\pm$ 0.0199 & 0.2194 $\pm$ 0.0267 \\
\gls{cem}-T2   & 22.66K & 11.08M & Matern             & \textbf{0.1623 $\pm$ 0.0170} & 0.2166 $\pm$ 0.0274 \\
\gls{cem}-T4   & 22.66K & 19.79M & Matern             & 0.1742 $\pm$ 0.0206 & 0.2168 $\pm$ 0.0271 \\
\gls{cem}-T8   & 22.66K & 37.20M & Matern             & 0.1676 $\pm$ 0.0174 & \textbf{0.2161 $\pm$ 0.0248} \\
\midrule
Plain    & 33.98K & 6.76M  & Periodic           & 0.0557 $\pm$ 0.0056 & 0.0614 $\pm$ 0.0058 \\
Gated    & 33.94K & 6.73M  & Periodic           & 0.0386 $\pm$ 0.0036 & 0.0445 $\pm$ 0.0043 \\
\gls{cem}-T1   & 22.66K & 6.73M  & Periodic           & 0.0352 $\pm$ 0.0036 & 0.0421 $\pm$ 0.0044 \\
\gls{cem}-T2   & 22.66K & 11.08M & Periodic           & 0.0240 $\pm$ 0.0022 & 0.0342 $\pm$ 0.0033 \\
\gls{cem}-T4   & 22.66K & 19.79M & Periodic           & \textbf{0.0238 $\pm$ 0.0029} & \textbf{0.0340 $\pm$ 0.0042} \\
\gls{cem}-T8   & 22.66K & 37.20M & Periodic           & 0.0242 $\pm$ 0.0037 & 0.0344 $\pm$ 0.0053 \\
\midrule
Plain    & 33.98K & 6.76M  & Rational Quadratic & 0.0791 $\pm$ 0.0109 & 0.0885 $\pm$ 0.0108 \\
Gated    & 33.94K & 6.73M  & Rational Quadratic & 0.0599 $\pm$ 0.0070 & 0.0715 $\pm$ 0.0075 \\
\gls{cem}-T1   & 22.66K & 6.73M  & Rational Quadratic & 0.0578 $\pm$ 0.0072 & 0.0709 $\pm$ 0.0082 \\
\gls{cem}-T2   & 22.66K & 11.08M & Rational Quadratic & 0.0409 $\pm$ 0.0039 & \textbf{0.0596 $\pm$ 0.0068} \\
\gls{cem}-T4   & 22.66K & 19.79M & Rational Quadratic & 0.0417 $\pm$ 0.0033 & 0.0606 $\pm$ 0.0067 \\
\gls{cem}-T8   & 22.66K & 37.20M & Rational Quadratic & \textbf{0.0406 $\pm$ 0.0035} & 0.0606 $\pm$ 0.0060 \\
\midrule
Plain    & 33.98K & 6.76M  & Non-Stationary     & 0.0385 $\pm$ 0.0052 & 0.0426 $\pm$ 0.0062 \\
Gated    & 33.94K & 6.73M  & Non-Stationary     & 0.0315 $\pm$ 0.0030 & 0.0358 $\pm$ 0.0039 \\
\gls{cem}-T1   & 22.66K & 6.73M  & Non-Stationary     & 0.0305 $\pm$ 0.0044 & 0.0356 $\pm$ 0.0051 \\
\gls{cem}-T2   & 22.66K & 11.08M & Non-Stationary     & 0.0257 $\pm$ 0.0027 & 0.0330 $\pm$ 0.0042 \\
\gls{cem}-T4   & 22.66K & 19.79M & Non-Stationary     & 0.0267 $\pm$ 0.0030 & 0.0330 $\pm$ 0.0043 \\
\gls{cem}-T8   & 22.66K & 37.20M & Non-Stationary     & \textbf{0.0243 $\pm$ 0.0030} & \textbf{0.0317 $\pm$ 0.0043} \\
\bottomrule
\end{tabular}
\end{table}

\begin{figure}[ht!]
\begin{center}
\includegraphics[width=0.5\textwidth]{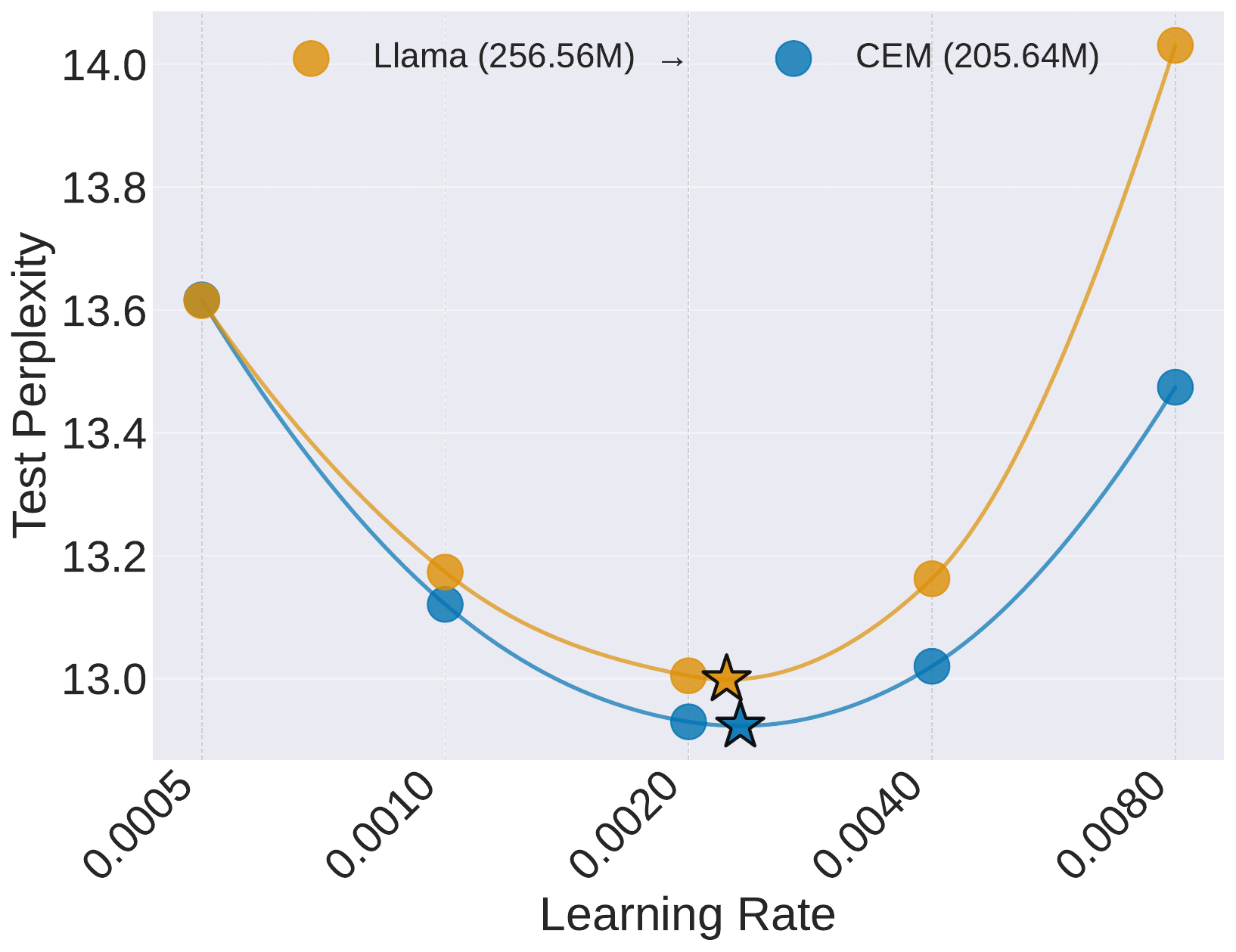}
\end{center}
\caption{Additional results for optimal learning-rate estimation via Akima interpolation.
Orange curves denote baseline Llama models; blue curves denote \gls{cem} models ($T=2$) with both MHA and \gls{mlp} replaced.
Star markers denote interpolated optima from five data points.
\gls{cem} models use roughly half the MHA parameters and two-third the \gls{mlp} parameters per layer, with two additional layers to offset the larger parameter reduction for this size, and are trained under the same token budget (Chinchilla-optimal for Llama).}  
\label{fig:akima_plot_256m}
\end{figure}

\textbf{Akima interpolation with larger models}
We scale our models to 256M parameters, and results analogous to \Cref{fig:akima} are shown in \Cref{fig:akima_plot_256m}. We find that CEM models, despite having fewer parameters, continue to outperform Llama models at this scale. Scaling to substantially larger sizes would require significantly more computational resources, and we leave this for future work.

\section{LLM Usage Statement} \label{appsec:llm_usage}

We used ChatGPT-5 to assist with paraphrasing, text editing, and proofreading. For most paragraphs, we first wrote a draft and then asked ChatGPT to paraphrase it without changing the original meaning. We checked that the paraphrased text did not alter the meaning.
We also used ChatGPT to help search for and discover relevant related work, but all bibliographic entries were manually verified for correctness. All conceptual development, technical contributions, experiments, and analysis were carried out by the authors.

\section{Reproducibility} \label{appsec:reproducibility}
We provide experimental details in \Cref{appsec:experimental_details}. Model architectures are given in \Cref{tab:model_configs}, and training configurations in \Cref{tab:training_config}. All experiments use the SlimPajama dataset (\Cref{appsec:data_preprocessing}) and were conducted on 8× NVIDIA A100 GPUs. The code is not yet publicly available but will be released upon publication.

\end{document}